\documentclass[]{interact}
\usepackage[colorlinks=true,linkcolor=blue,citecolor=blue,urlcolor=blue]{hyperref}

\usepackage{url}
\usepackage{epstopdf}
\usepackage{subfigure}

\usepackage{natbib}
\bibpunct[, ]{(}{)}{,}{a}{}{,}
\renewcommand\bibfont{\fontsize{10}{12}\selectfont}

\theoremstyle{plain}

\theoremstyle{definition}

\theoremstyle{remark}

\usepackage{xcolor}

\begin{document}


\title{Subnational Geocoding of Global Disasters Using Large Language Models}

\author{
\name{Michele Ronco\textsuperscript{a}\thanks{Email: michele.ronco@ec.europa.eu}, Damien Delforge\textsuperscript{b}, Wiebke S. Jäger\textsuperscript{c}, Christina Corbane\textsuperscript{a}}
\affil{\textsuperscript{a}European Commission, Joint Research Centre, Ispra, 21027, Italy; \textsuperscript{b}Institute of Health and Society (IRSS), University of Louvain (UCLouvain), Clos Chapelle-aux-champs 30, Woluwe St Lambert, 1200, Brussels, Belgium; \textsuperscript{c}Institute for Environmental Studies, Vrije Universiteit Amsterdam, Amsterdam, the Netherlands}
}

\maketitle

\begin{abstract}
Subnational location data of disaster events are critical for risk assessment and disaster risk reduction. Disaster databases such as EM-DAT often report locations in unstructured textual form, with inconsistent granularity or spelling, that make it difficult to integrate with spatial datasets. We present a fully automated LLM-assisted workflow that processes and cleans textual location information using GPT-4o, and assigns geometries by cross-checking three independent geoinformation repositories: GADM, OpenStreetMap and Wikidata. Based on the agreement and availability of these sources, we assign a reliability score to each location while generating subnational geometries. Applied to the EM-DAT dataset from 2000 to 2024, the workflow geocodes 14,215 events across 17,948 unique locations. Unlike previous methods, our approach requires no manual intervention, covers all disaster types, enables cross-verification across multiple sources, and allows flexible remapping to preferred frameworks. Beyond the dataset, we demonstrate the potential of LLMs to extract and structure geographic information from unstructured text, offering a scalable and reliable method for related analyses. 
\end{abstract}


\section{Introduction}

Accurate and standardized disaster data form the foundation of effective risk assessment, and disaster risk reduction. Global frameworks such as the Sendai Framework for Disaster Risk Reduction (SFDRR) emphasize the need for evidence-based approaches that account for the full spectrum of hazards — natural, technological, biological, and societal — and their compounding impacts \citep{UN15}. Yet, operationalizing disaster information remains challenging: global databases such as the Emergency Events Database (EM-DAT), which provide long-term and wide-ranging coverage, often contain incomplete, inconsistent, or ambiguous location data \citep{Del25}. Location fields contain unstructured text with mixed granularity, inconsistent spellings, and homonyms, while records rarely specify the administrative reference system. These limitations are especially critical in the context of multi-hazard events, where simultaneous or cascading processes produce disproportionate impacts. A substantial share of global economic losses stems from such compound interactions (\cite{Zsc17}, \cite{Zsc20}, \cite{Zsc24}, \cite{Jag25}, \cite{Ron25}), yet most databases retain a single-hazard lens. Addressing these gaps requires methods that can consistently assign geometries to disaster events, enabling spatial analyses of compound and multi-hazard interactions.

Recent advances in large language models (LLMs), a class of artificial intelligence (AI) foundation models, have created a versatile framework capable of supporting a wide range of applications \citep{Cha24}, including geospatial reasoning \citep{Jan25}. Trained on extensive text corpora, LLMs implicitly acquire knowledge about places, administrative hierarchies, naming conventions across scales, and spatial relationships among locations, enabling them to interpret and structure geographic information ( \cite{Man23}, \cite{Zha24}, \cite{Hu24}). In this sense, models such as GPT can be viewed as ``geographers in AI form". Moreover, LLMs can automate repetitive and time-consuming geocoding and data-cleaning tasks that traditionally require substantial manual effort, offering new opportunities for scalable and consistent disaster data processing.

Building on these capabilities, we introduce an automated, LLM-assisted workflow for geocoding EM-DAT disaster records from 2000 to 2024, encompassing 14,215 events across 17,948 locations. To our knowledge, this approach is the first to use LLMs for subnational geocoding of global disaster data. Using GPT-4o, we transform raw location descriptions into hierarchical JSON representations, which are then geocoded into polygon or point geometries by cross-checking three independent geoinformation sources: OpenStreetMap (OSM), Wikidata, and Global Administrative Areas (GADM). Based on the agreement and availability of these sources, the system assigns a reliability score to each event. The workflow operates end-to-end with minimal manual intervention and accommodates all disaster types. To assess the performance of the LLM-based geocoding, we compare a subset of events with two independent reference datasets—the manually geocoded EM-DAT locations \citep{Del24} and the Geocoded Disasters (GDIS) dataset \citep{Rosvold21}. Our contributions include:

\begin{enumerate}
    \item A fully automated LLM-assisted geocoding workflow that integrates multiple geospatial knowledge bases for robust subnational location identification.
    \item LLM-GeoDis: a global, openly available dataset of geocoded disaster events, complete with spatial confidence estimates.
    \item An empirical evaluation of GPT’s geographic and toponymic reasoning through comparison with two independent reference datasets, demonstrating consistent performance across world regions with no evident bias toward better-documented or high-income areas.
\end{enumerate}

The remainder of this paper is organized as follows. Section 2 reviews related work on disaster geocoding and previous efforts to enhance the spatial information in EM-DAT and other global disaster-related datasets. Section 3 describes our LLM-assisted geocoding methodology and workflow, including data processing, triangulation across geospatial sources, and confidence estimation. Section 4 presents the resulting geocoded dataset, along comparison with existing geocoding methods. Section 5 discusses the main findings, applications, and limitations of the approach, and concludes with perspectives for future research.

\section{Related work}

Previous efforts to geocode EM-DAT records have focused on linking disaster events to georeferenced polygons or points. The EM-DAT project initiated the geocoding of disaster events in 2013, with a retrospective analysis dating back to 2000 \citep{Del25}. They manually geocoded only natural-hazard disasters, excluding biological hazards, using online map services and the Food and Agriculture Organization Global Administrative Unit Layer (GAUL, version 2015). The geocoded administrative units have been publicly accessible since 2020 from the EM-DAT data portal (\href{https://public.emdat.be/}{}). Overall, the geocoding coverage is $54\%$ for the 2000-2023 period \citep{Del24}.

Another initiative, independent of EM-DAT, is the Geocoded Disasters (GDIS) dataset \citep{Rosvold21}. GDIS introduces a semi-automated approach based on EM-DAT’s location string matching to GADM units, combined with manual reconciliation to assign unmatched locations. GDIS is limited to events recorded up to 2018 and relies on manual intervention for approximately half of the locations, and does not provide uncertainty estimates for ambiguous locations. GDIS is also limited to floods, storms, earthquakes, volcanic activity, extreme temperatures, landslides, droughts, and (dry) mass movements (see also Table \ref{tab:comparison}). The more recent Geo-Disasters framework covers  1990–2023 period and completes the initial EM-DAT GAUL geocoding primarily for the 1990-1999 period, using GeoNames queries, string matching, and manual corrections for the remaining locations, with matching quality flags \citep{Teber25}.

Other studies have addressed the geocoding of disaster-induced human displacement events, which present challenges similar to those encountered in EM-DAT. \cite{Ronco23} compiled a global database of displacement events from floods, storms, and landslides (2016–2021), geocoding affected areas using OpenStreetMap. When no precise polygon of the affected area was available, they assigned the event to the smallest corresponding administrative unit that could be matched by name. \cite{Mester23} focused specifically on flood-induced displacement, using string similarity matching to GADM administrative units. While both approaches enabled subnational analysis, they relied on simple rule-based or string-matching criteria rather than AI-based methods. As a result, a substantial fraction of event locations remained unmatched or only approximately identified, limiting the completeness and spatial accuracy.

\section{Methodology }

\subsection{Data}

EM-DAT is a global database that compiles data on the human and economic impacts of disasters from natural and technological hazards since 1900, serving the humanitarian, disaster risk reduction, and academic sectors \citep{Del25}. Here we use three EM-DAT’s reported fields, (i) a unique disaster identifier (\texttt{DisNo.}), (ii) the \texttt{Location} column, as a free-text field, reporting the impacted locations mentioned in the source textual document, and (iii) the \texttt{Country} column referring to the impacted countries. The \texttt{Location} column contains no standard formatting or consistent separator and may list locations in any order or at varying levels of granularity (e.g., ``country'', ``region'', ``municipality'', ``village''), often including parenthetical clarifications, multiple languages, or minor typos. 

We focus on the years 2000–2024 to avoid known biases in historical reporting prior to 2000 \citep{Del25}. Over this period, the original EM-DAT dataset contains 15,404 records with a non-null \texttt{Location} field. Within the full dataset, 8,402 events (about $54\%$) also include manually assigned administrative units provided by EM-DAT. We intentionally do not use these manual geocodes in our workflow, relying solely on the original free-text \texttt{Location} field to ensure an independent, fully automated geocoding process. However, the EM-DAT administrative unit labels are later used as an external reference for evaluating our results (see Section 4).

\subsection{Geocoding process}

\begin{figure}[h!]
\centering
\includegraphics[width=12cm,height=15cm]{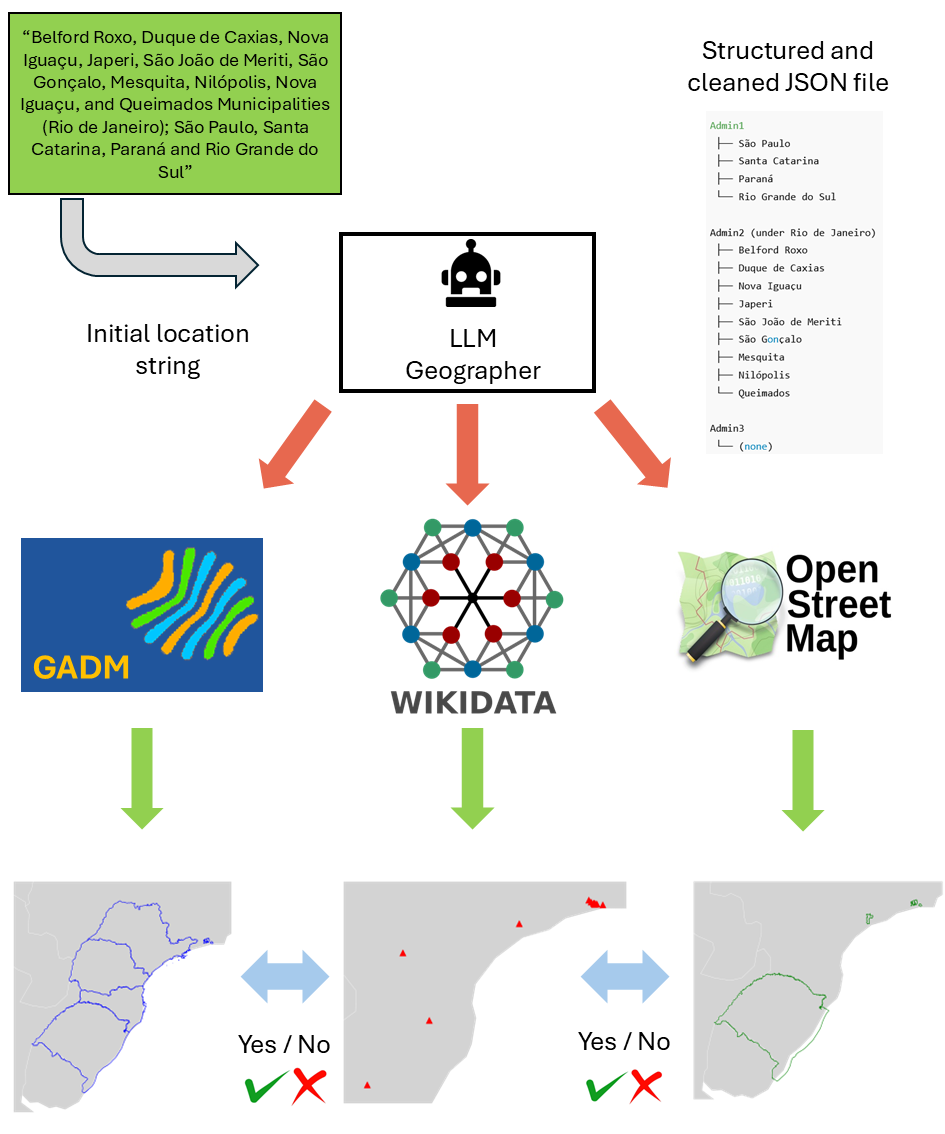}
\caption{Overview of the geocoding pipeline. Free-text location strings from EM-DAT are first parsed by GPT-4o, which outputs a structured JSON representation of the identified hierarchical administrative units. This intermediate representation is then processed independently through three geocoding procedures—GADM, OpenStreetMap, and Wikidata—to obtain candidate geometries. The resulting geometries are cross-checked across sources to establish consistency and confidence in the final spatial representation of each disaster location.}
\label{fig:pipeline}
\end{figure}

Building on previous efforts to geocode disaster records (see Section 2), this study develops a geocoding pipeline for transforming free-text location descriptions from the EM-DAT disaster database into standardized, spatially explicit geometries. The overall workflow is illustrated in Figure \ref{fig:pipeline}.

At a high level, the geocoding pipeline consists of the following steps:

\begin{enumerate}
    \item LLM-based parsing of free-text locations.
Free-text location strings from EM-DAT are processed with GPT-4o, which extracts and structures place references into a hierarchical JSON representation. 
    \item Independent geocoding using three spatial reference sources.
The structured JSON output is submitted to three parallel geocoding procedures—GADM, OpenStreetMap, and Wikidata—each producing candidate geometries. 
    \item Cross-source consistency and reconciliation.
Candidate geometries from the three sources are compared to assess consistency, resolve discrepancies, and determine reliability.
    \item Final location assignment (GADM-based harmonization).
All candidate geometries are reconciled and reprojected onto the GADM administrative hierarchy, which serves as the unified spatial reference system. 
\end{enumerate}

\subsubsection{Parsing}

In the first step, we employ GPT-4o to preprocess the raw location strings into structured and consistent hierarchical JSON data. GPT-4o, developed by OpenAI, is a state-of-the-art large language model (LLM) with around 200 billion parameters trained on diverse multilingual text corpora that include geographic and spatial information. By leveraging its learned knowledge of place names, their relationships, and spatial contexts, in our workflow GPT-4o functions as a \textit{geographer}, interpreting and structuring unstandardized or ambiguous text through few-shot or zero-shot learning.

\begin{figure}[h!]
\centering
\includegraphics[width=12cm,height=16cm]{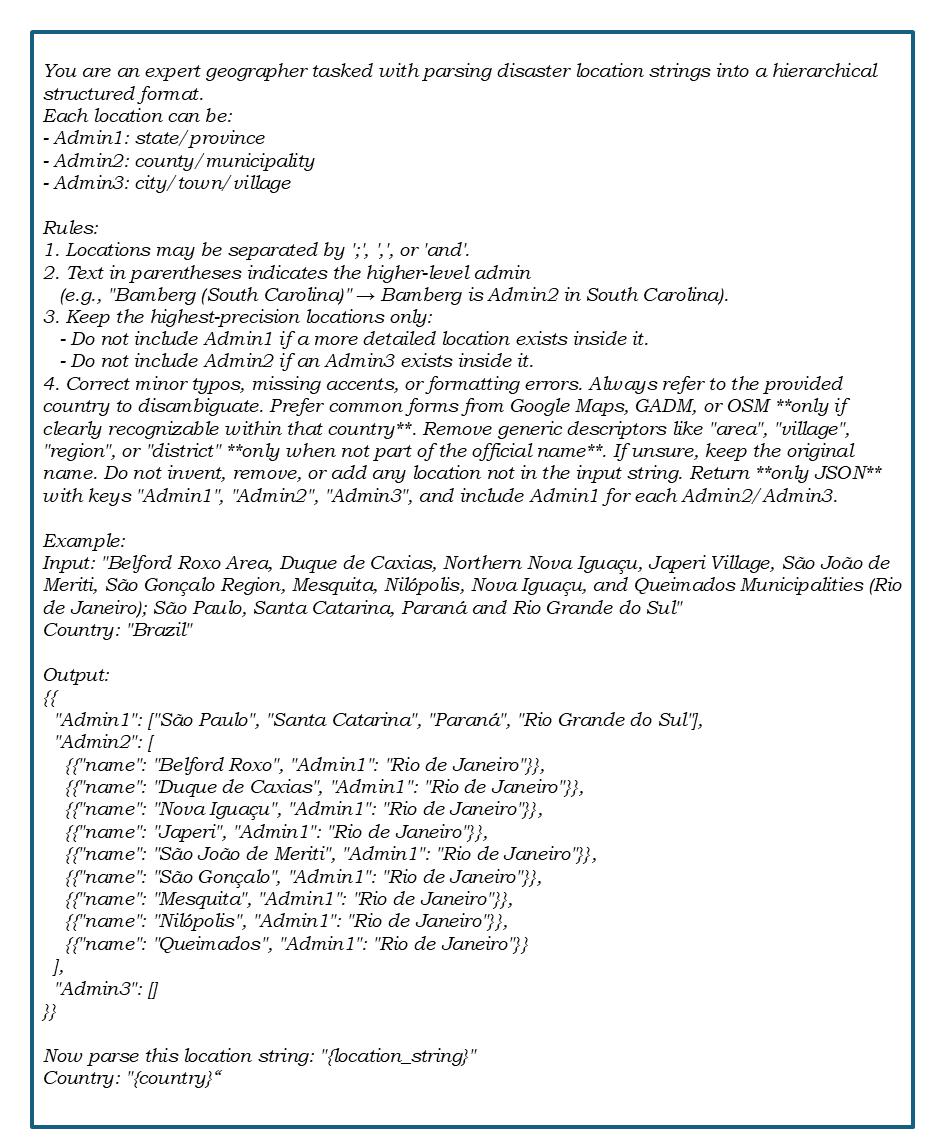}
\caption{GPT-4o prompt used for parsing disaster location strings with ICL. The prompt provides a sample input and the expected JSON output to guide the model in identifying unique locations, resolving granularity, cleaning names, and organizing them hierarchically into \texttt{Admin1}, \texttt{Admin2}, and \texttt{Admin3} levels. The name of the country is provided to reduce toponymic ambiguities. This prompt ensures that GPT-4o produces a structured canonical representation suitable for subsequent geocoding.}
\label{fig:prompt}
\end{figure}

Using in-context learning (ICL), the model is provided with an example location string and the corresponding JSON to guide parsing (see Figure \ref{fig:prompt} for the full prompt). The prompt instructs GPT-4o to identify and clean individual place names, correct minor typographical and formatting errors, and determine the administrative granularity of each location. The model then arranges these entities into a hierarchical structure comprising up to three administrative levels. Parenthetical expressions are interpreted as higher-level administrative units, and lower-level entities are linked to their corresponding \texttt{Admin1} or \texttt{Admin2}. The \texttt{Country} field is provided as contextual information to guide interpretation and disambiguation—particularly in cases where administrative names are shared across countries. Generic descriptors such as ``area'', ``village'', or ``district'' are removed when they are not part of the official toponym. The output JSON contains only the keys \texttt{Admin1}, \texttt{Admin2}, and \texttt{Admin3}, with multiple place names represented appropriately. Each output record follows a standardized format, where \texttt{Admin1}, \texttt{Admin2}, and \texttt{Admin3} correspond to progressively finer administrative levels; lower-level entries include their parent administrative unit, ensuring that hierarchical relationships are explicitly preserved.

The JSON-structured data produced by GPT-4o serves as the canonical representation of each disaster location, broken down by separate locations mentioned in the string, to support subsequent matching against multiple geospatial data sources. By decoupling parsing from geocoding, the approach allows the same structured representation to be reused across independent geocoding procedures. To ensure robustness, parsing implements retries with exponential backoff in case of temporary service interruptions. Additionally, all results are cached locally to avoid re-geocoding locations that appear multiple times, thereby reducing computation and improving the efficiency of the procedure.

\subsubsection{Geocoding}

In the second step, the structured JSON data is processed through three independent geocoding workflows, each applying a different strategy to assign coordinates or polygons to the parsed locations. For the OSM Nominatim and Wikidata APIs, the querying process incorporates rate-limiting and a retry mechanism with short randomized backoff intervals to handle temporary connection failures or rate-limit responses. GADM, by contrast, is processed locally from a downloaded dataset and does not require API-based requests. Before geocoding, all location names are standardized through a normalization step that lowercases text, removes accents and parenthetical content, strips generic descriptors (e.g., ``region'', ``district'', ``province''), and cleans formatting inconsistencies such as redundant whitespace or punctuation.

The first workflow leverages the GADM v4.1 dataset to match parsed locations to standardized administrative units. Matching is performed hierarchically: \texttt{Admin1} units are matched first, followed by \texttt{Admin2} units constrained to their respective parent \texttt{Admin1}. A fuzzy string similarity approach (rapidfuzz.process.extractOne with the WRatio scorer) identifies the best match for each location \citep{Bac25}. WRatio combines edit-distance, substring, and token-based comparisons into a single similarity score (0–100), making it robust to typos, word-order differences, and minor variations. Only matches exceeding a predefined threshold ($85\%$) are retained. 

Second, place names are geocoded using OpenStreetMap via the Nominatim API. The hierarchical JSON structure is used to build context-aware queries by combining each location with its parent administrative units (e.g., $[\texttt{Admin2}, \texttt{Admin1}]$ or $[\texttt{Admin1}, \texttt{Country}]$). Including this hierarchical context improves spatial precision and helps disambiguate places that share the same name, enabling more accurate retrieval of point or polygon geometries. 

The third workflow queries Wikidata via SPARQL to identify candidate entities for each location. For each hierarchical level (\texttt{Admin1}, \texttt{Admin2}, \texttt{Admin3}), the system constructs a SPARQL query to retrieve candidate entities matching the place name, along with their coordinates, country labels, and unique Wikidata identifiers (QIDs). Candidate entities are parsed and ranked based on country consistency and spatial proximity to parent administrative units when applicable. When a parent geometry (e.g., \texttt{Admin1} for an \texttt{Admin2} candidate) is available, geodesic distances are computed and the closest consistent candidate is selected. Otherwise, the top-ranked match within the target country is retained. 

The outputs from all three workflows are stored in a structured hierarchical dictionary that records the matched names and their associated geometries while preserving parent–child administrative relationships.

\subsubsection{Consistency among sources}

To evaluate the spatial consistency of geocoded records, we developed a matching reliability score, a composite indicator that quantifies the agreement and completeness of spatial information across three geographic datasets: GADM, OSM, and Wikidata. The score incorporates (i) the presence of relevant geometries in each dataset, (ii) the percentage of spatial overlap between GADM and OSM boundaries, and (iii) the containment of Wikipedia geometries within GADM and OSM areas. The metric ranges from 0 to 4, where higher scores indicate stronger inter-dataset concordance and thus greater confidence in the assigned location. Importantly, while locations confirmed by multiple sources receive higher reliability, those supported by a single dataset may still be correct. This approach extends previous studies (\cite{Rosvold21}, \cite{Teber25}), which typically relied on a single geospatial reference, by providing a more nuanced and transparent assessment of spatial reliability.

\subsubsection{Re-projection to GADM}

After obtaining candidate geometries from GADM, OSM, and Wikidata, all locations are harmonized onto the GADM administrative framework. For records lacking a direct GADM match, geometries from OSM or Wikidata—preferring the latter when available—are used as proxies. Each proxy geometry is intersected with all GADM administrative units, and the area of overlap is computed to identify the most likely corresponding region. When intersection is found, the location is reassigned to the GADM unit with the largest overlap, and the missing administrative name and boundary are filled accordingly. This spatial-overlap procedure ensures that all locations are ultimately expressed using a consistent GADM-based representation. A similar overlap-based remapping approach can be applied to project the geocoded results onto any other desired reference administrative system.

\begin{figure}[h!]
\centering
\includegraphics[width=15cm,height=9cm]{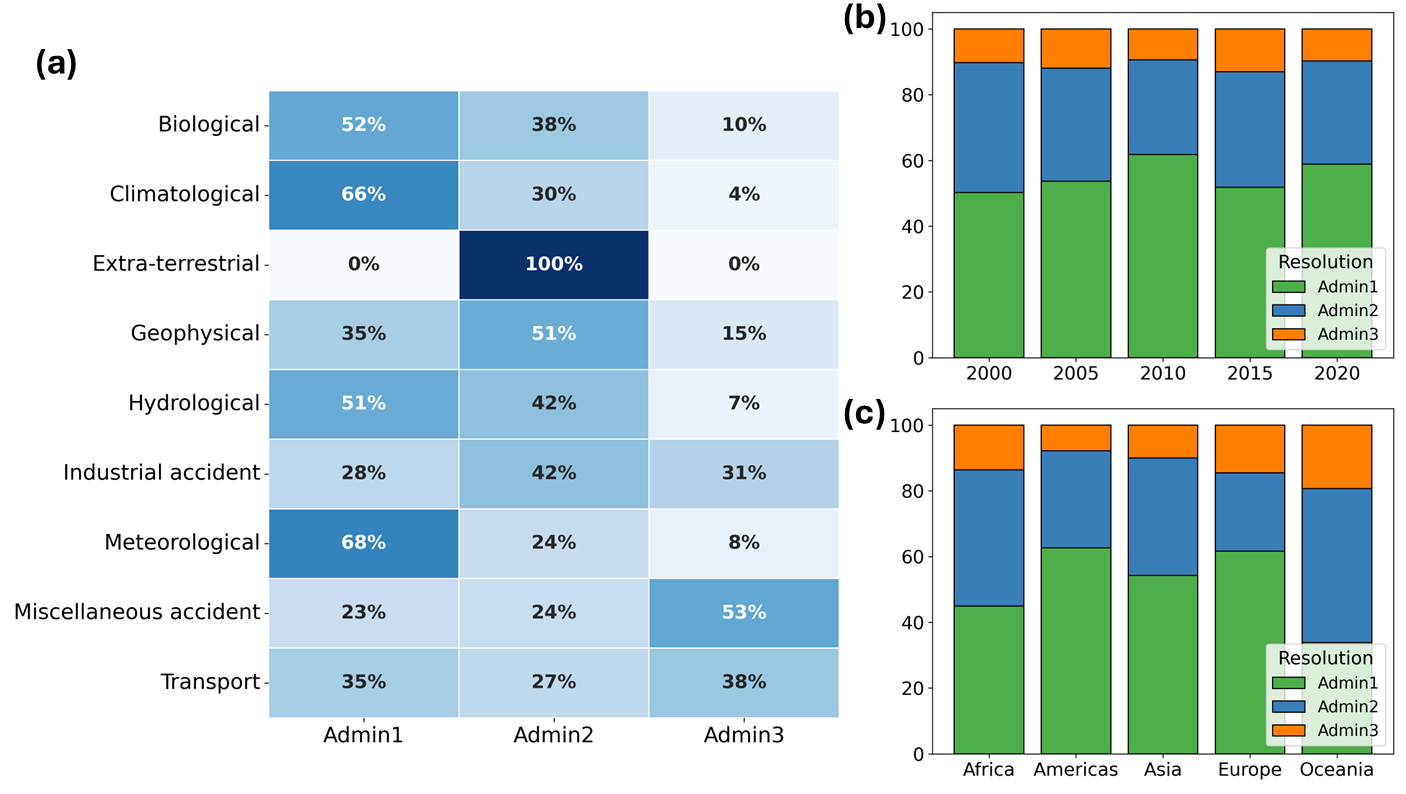}
\caption{Geocoding coverage and resolution of disaster records (2000–2024). (a) Heatmap showing the proportion of events geocoded by disaster subgroup. (b) Stacked bar chart of geocoded records every five years, illustrating temporal trends. (c) Regional distribution of geocoded disasters, showing relative contributions across continents.}
\label{fig:admins}
\end{figure}

\section{Results}

\subsection{Analysing LLM-GeoDis}

Between 2000 and 2024, our workflow successfully geocoded 14,215 disaster records from the EM-DAT database, resulting in the LLM-GeoDis dataset. This corresponds to $92\%$ of all entries containing some location information and $88\%$ of all recorded disasters within this period. These events span nine disaster subgroups and 31 disaster types, following EM-DAT’s classification scheme. In total, the geocoded dataset comprises 17,948 unique location entries. The resulting dataset assigns $55.27\%$  of the records at \texttt{Admin1}, $33.84\%$  of the records at \texttt{Admin2}, and $10.89\%$  of the records at \texttt{Admin3}, reflecting the distribution of spatial detail recovered by the geocoding pipeline. 

\begin{figure}[h!]
\centering
\includegraphics[width=15cm,height=9cm]{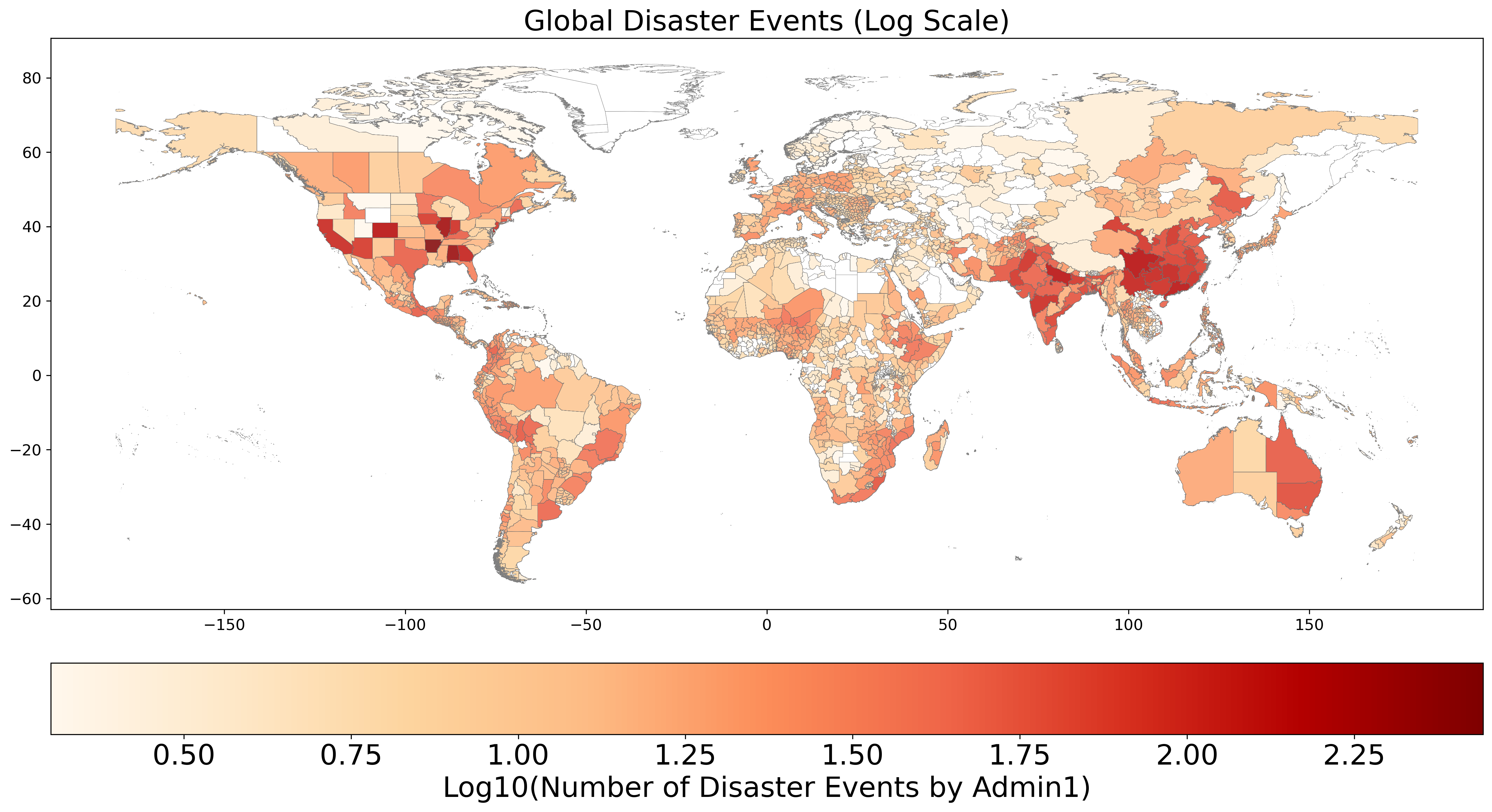}
\caption{Distribution of reported Disaster Events (2000–2024). This map displays disaster occurrences aggregated at the GADM \texttt{Admin1} level, providing a visual overview of event frequency across regions. The original data, available at a finer resolution, has been remapped to \texttt{Admin1} for consistency and clarity in global analysis.}
\label{fig:globmap}
\end{figure}

Figure \ref{fig:admins} summarizes the granularity and coverage of geocoded disaster records across disaster subgroups, time, and regions. Panel (a) highlights marked differences in administrative levels among disaster types: climatological and meteorological disasters are predominantly reported at the \texttt{Admin1} level (around two-thirds of records), while industrial, transport, and miscellaneous accidents show a much higher share of \texttt{Admin3}-level reporting. Geophysical and hydrological disasters display a more balanced distribution between \texttt{Admin1} and \texttt{Admin2}, with geophysical events including a notable proportion at \texttt{Admin3}. Panel (b) illustrates the temporal evolution of spatial detail: over the two-decade period, \texttt{Admin1} consistently accounts for the largest share of records. \texttt{Admin2} representation decreases from $39\%$ in 2000 to $31\%$ in 2020, while \texttt{Admin3} remains relatively stable but consistently minor throughout. Panel (c) compares regional patterns, showing that Africa maintains a balanced distribution between \texttt{Admin1} and \texttt{Admin2}, the Americas and Europe emphasize \texttt{Admin1}-level reporting, and Oceania exhibits the highest proportion of \texttt{Admin3} entries.

Figure \ref{fig:globmap} presents the spatial distribution of disaster occurrences between 2000 and 2024, aggregated at the GADM \texttt{Admin1} level. The map reveals clear spatial clustering of disaster activity across major regions. The United States, China, and India record the highest number of subnational occurrences, reflecting both their geographic extent and exposure to multiple hazard types. Within the United States, several central and southeastern states—such as Arkansas, Alabama, and Illinois—show particularly high event frequencies. In China, disaster occurrences are concentrated in Sichuan, Guangdong, and Chongqing, while in India, high counts are observed in Uttar Pradesh, Maharashtra, and Bihar. Other recurrent hotspots include Dhaka (Bangladesh), Punjab and Sindh (Pakistan), and Aurora and Albay (Philippines), corresponding to regions regularly affected by floods, cyclones, or seismic activity. For more insight on hotspots of climate-related disasters in EM-DAT, we refer to \citep{Don24}.

Applying the matching reliability scoring framework revealed a high degree of spatial agreement across the three datasets. On average, the GADM–OSM boundary overlap was $86.5\%$, while $89.2\%$ of Wikipedia geometries were contained within GADM and $85.8\%$ within OSM boundaries. These results indicate strong cross-dataset concordance and support the robustness of the geocoding process. Locations confirmed by multiple sources were classified as high reliability, whereas those supported by only one dataset were labeled low reliability, reflecting lower cross-dataset agreement rather than necessarily incorrect geocoding. The reliability score thus provides a relative measure of confidence grounded in observed inter-dataset consistency.

\subsection{Comparative analysis} 

\begin{table}[h!]
\tbl{Comparison of datasets: LLM-GeoDis, EM-DAT GAUL, and GDIS}{
\begin{tabular}{lp{3cm}p{3cm}p{3cm}}
\toprule
 & LLM-GeoDis & EM-DAT GAUL & GDIS \\ 
\midrule
Time period & 2000--2024 & 2000--2023 & 1960--2018 \\ 
Disaster types & All types & All types & {\raggedright Floods, storms, earthquakes, volcanic activity, extreme temperatures, landslides, droughts, and (dry) mass movements.} \\ 
ADMIN reference & Wikidata–OSM–GADM & GAUL & GADM \\ 
Resolution & Up to Admin3 & Up to Admin2 & Up to Admin3 \\ 
Total records & 14,215 & 8,399 & 9,018 \\ 
Unique locations & 17,948 & 10,635 & 10,617 \\ 
\bottomrule
\end{tabular}}
\label{tab:comparison}
\end{table}

To evaluate our geocoding algorithm, we compare the geometries produced by LLM-GeoDis at (GADM, OSM, and Wikidata) with manually geocoded geometries in the EM-DAT 1900-2023 Archive \citep{Del24}. Admin 3 units from GADM and OSM are converted into Admin 2 units to match the administrative level of EM-DAT GAUL geometries. Additionally, we compare geometries from other geocoding efforts, such as GDIS \citep{Rosvold21}, with those from EM-DAT for further insights. We identify and select the disaster numbers that are present in both datasets. 

\begin{enumerate}
    \item We conduct comparative analyses on two levels: (i) the individual administrative units’ level and (ii) the disaster footprint level, where the geometries of administrative units linked to the same disaster are merged. In both cases, the individual or merged geometries are compared to the disaster footprint created by merging the benchmark geometries.   
    \item We report three metrics to compare geometries. Suppose A is a geometry in one dataset and B is a geometry in the benchmark dataset:
    \begin{itemize}
        \item A in B: the ratio (0-1) of the area of A that falls within B, or  $\frac{A \cap B}{A}$.
        \item B in A: the ratio (0-1) of the area of B that falls within A, or  $\frac{A \cap B}{B}$. 
        \item  The Jaccard index: a standard similarity indicator (0-1), representing the ratio of the intersection area to the union area, or $J(A, B) = \frac{|A \cap B|}{|A \cup B|}
$.
    \end{itemize}
\end{enumerate}

For LLM-GeoDis Wikidata point geometries, A in B is reported differently, using a binary containment index (0, 1) that indicates whether the point falls within the benchmark geometry. 
Some key features of the three datasets we compare are reported in Table \ref{tab:comparison}

\begin{figure}[h!]
\centering
\includegraphics[width=14cm,height=7cm]{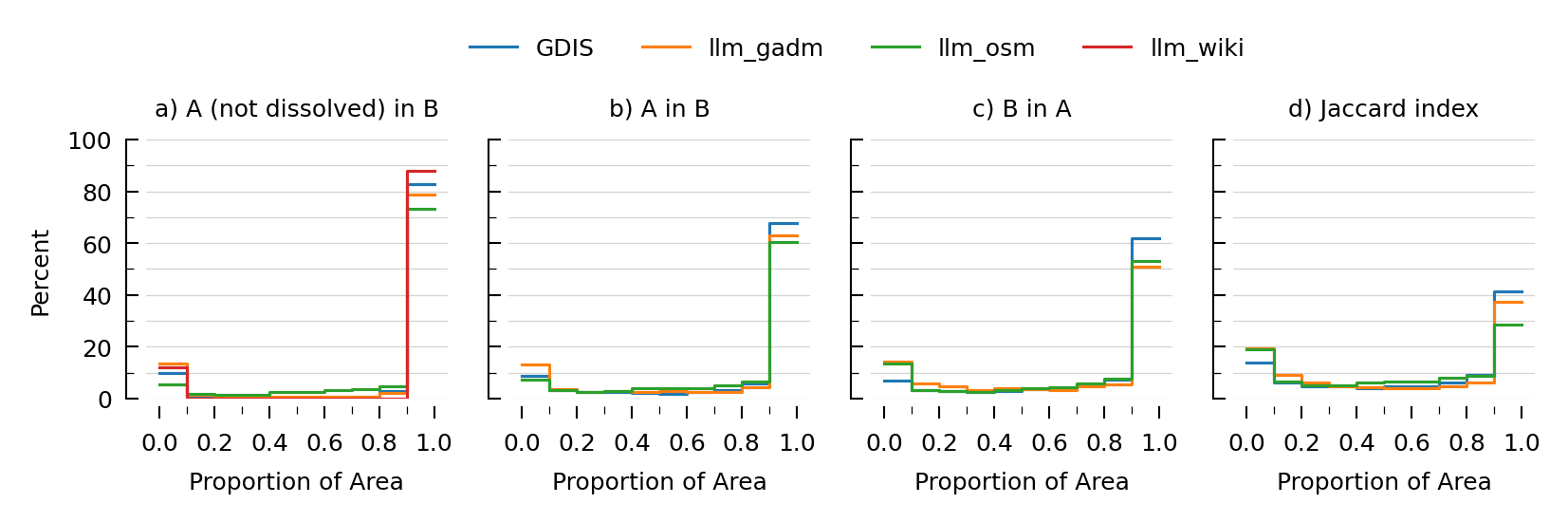}
\caption{Area overlap of candidate geometries from datasets derived with (semi-)automatic geocoding (GDIS, LLM-GADM, LLM-OSM, and LLM-Wiki) with benchmark geometries (EM-DAT GAUL). The panels show histograms for different overlap metrics: a) shows the percentage of proportions of single-location administrative candidate area included in the benchmark disaster area. The LLM-Wiki data with Point geometries are reported as a binary variable (i.e., 0 for non-inclusion, and 1 for inclusion). b) shows the percentage of proportions of the candidate disaster area (i.e., dissolved single-location geometries sharing the same DisNo.) included in the benchmark disaster area; c) shows the percentage of proportions of the benchmark disaster area included in the candidate disaster area; and d) shows the percentage of the Jaccard index between the candidate disaster areas and the benchmark disaster areas.
}
\label{fig:val}
\end{figure}

Figure \ref{fig:val} summarizes the results of the comparative analysis using the manually geocoded EM-DAT records as a benchmark for disaster footprints. Panel (a) enables checking whether individual geocoded units are included in the EM-DAT manually geocoded footprints. The other three panels (b-d) compare geometries at the impact footprint level, whether they are included in each other (A in B, B in A) or overlap in a similar manner (Jaccard index). Each plot reports the distribution of metrics as percentages, primarily U-shaped and skewed toward strong agreement (ratio of 1), with a bin size of 1 decimal point. The point geometries of LLMGeoDis (Wiki) are benchmarked using a binary variable (1 if contained) in panel (a) only, since they cannot be dissolved into disaster footprint geometries.

In all cases, more than $70\%$ of individual units—almost $90\%$ for the Wikidata point geometries—have more than $90\%$ of their area falling within the EM-DAT GAUL footprint (Fig. 5a). The median values of the ratios are high: 0.990 (GDIS), 0.992 (GADM), and 0.990 (OSM). The mean ratios are naturally lower: 0.862 (GDIS), 0.823 (GADM), and 0.850 (OSM). Notably, the [0-0.1] bin includes both non-intersecting and barely intersecting units, which warrant distinction. The percentage of non-intersecting units for panel a is $5.64\%$ for GDIS, which is approximately half of the units in the [0,0.1] bin. For LLMGeoDis, $2.12\%$ (GADM), $2.39\%$ (OSM), and $11.99\%$ (Wiki). Hence, $97.9\%$ of the LLMGeoDis GADM individual geocoded units intersect the manually geocoded disaster footprints in EM-DAT. 

When examining footprint level statistics (b to d), the variability in overlap ratios increases, underscoring that disaster footprints are either not consistently geocoded at the same administrative levels and spatial resolutions, that administrative units may have undergone changes over time and between reference versions (Table 1), or that these geocoding workflows have distinct biases and systematic errors, which are challenging to estimate given that the EM-DAT GAUL benchmark may itself be prone to manual geocoding errors. Nonetheless, GDIS shows the strongest agreement with EM-DAT GAUL, with a $40\%$ proportion—slightly higher than LLMGeoDis GADM— of Jaccard indices above 0.9. The median Jaccard values are 0.804 (GDIS), 0.658 (GADM), and 0.629 (OSM). The mean Jaccard values are 0.644 (GDIS), 0.568 (GADM), and 0.560 (OSM). Lastly, the frequency of non-intersecting disaster footprints is $1.25\%$ (GDIS), $1.04\%$ (GADM), and $1.13\%$ (OSM).

\section{Discussion}

Accurate subnational geocoding is essential for disaster risk assessment, exposure modeling, and policy monitoring under frameworks such as the SFDRR and SDGs, which call for more disaggregated data \citep{Cla18}. Comprehensive access to georeferenced footprints further enables data interoperability with other geospatial datasets to build up additional analytical insights and risk awareness (\cite{Ronco23}, \cite{Mester23}, \cite{Mac25}, \cite{Ron25}). Although EM-DAT provides subnational location information, coverage is incomplete and references are often outdated, leaving many event locations ambiguous for spatial modeling.

Our workflow addresses the limitations of existing disaster databases by using GPT-assisted parsing to convert free-text location descriptions into structured administrative hierarchies and by cross-checking them against GADM, OpenStreetMap, and Wikidata. This process yields up to three candidate geometries per location and assesses their agreement, producing a reliability score: high scores indicate consensus across sources, while lower scores reflect reduced agreement but may still be correct. By operating on full administrative polygons rather than centroids or single-point proxies, the approach more accurately represents multi-region and areal hazards and is applicable to all disaster types, including complex cascading events. Although LLMGeoDis shows slightly lower alignment with EM-DAT GAUL than GDIS, it produces substantially fewer non-overlapping units, reflecting additional constraints applied to reduce ambiguity in subnational assignments. Its higher agreement with GADM compared to OSM or Wikidata workflows highlights the benefit of integrating multiple sources for more consistent geocoding. The workflow can be extended to future periods or other event databases such as DesInventar \citep{Pan20} or Internal Displacement Updates \citep{Ronco23}. AI-assisted parsing reduces manual workloads, enhances efficiency, and produces reproducible outputs suitable for timely disaster monitoring. Beyond methodological gains, subnational geocoding has clear policy relevance: accurate administrative assignments enable integration with hazard, population, and infrastructure layers, supporting exposure assessment, multi-hazard risk modeling, and decision-making under global frameworks.

Limitations at the source remain. EM-DAT location fields are extracted from reports or news, which may be incomplete, inconsistent, or ambiguous. While LLMs can assist in parsing, structuring, and disambiguating locations, maintaining a human-in-the-loop approach is essential to ensure data quality and resolve complex or uncertain cases. Future work could explore more extensive use of AI, for example by combining models for raw location extraction and hierarchical structuring, or by employing retrieval-augmented generation (RAG) to cross-reference news, web searches, or alternative databases. Integrating these capabilities into an end-to-end framework could further enhance the efficiency, consistency, and timeliness of disaster location, while human oversight ensures reliability, particularly for high-impact events. By providing harmonized subnational geocodes, this workflow also represents a step toward greater intercomparability across disaster datasets, allowing more reliable assessments of impacts at finer spatial scales rather than relying solely on country-level aggregates.

\section*{Acknowledgement(s)}

MR thanks Lorenzo Bertolini for valuable early-stage discussions on large language models and prompting approaches. MR also thanks Samuel Roeslin, Sandro Salari, Melina London, and Filippo Maria D'Arcangelo for revising and validating an early version of LLM-GeoDis, and Michele Melchiorri for his support and insightful discussions on the applications of geocoded disaster data. The authors thank Valentin Wathelet and Regina Below for their sustained efforts in manually geocoding disaster data as part of the EM-DAT project.

\section*{Data and code availability}

The LLM-GeoDis dataset is available at \url{https://doi.org/10.5281/zenodo.17544932}. All the codes to reproduce the analysis in this paper are available at \url{https://github.com/em-dat/emdat_geocoding}.

\section*{Disclosure statement}

The authors declare no conflict of interest.

\section{Appendices}

This appendix presents additional analyses that complement the main results.

\appendix

\section{Location distribution}

\begin{figure}[h!]
\centering
\includegraphics[width=15cm,height=10cm]{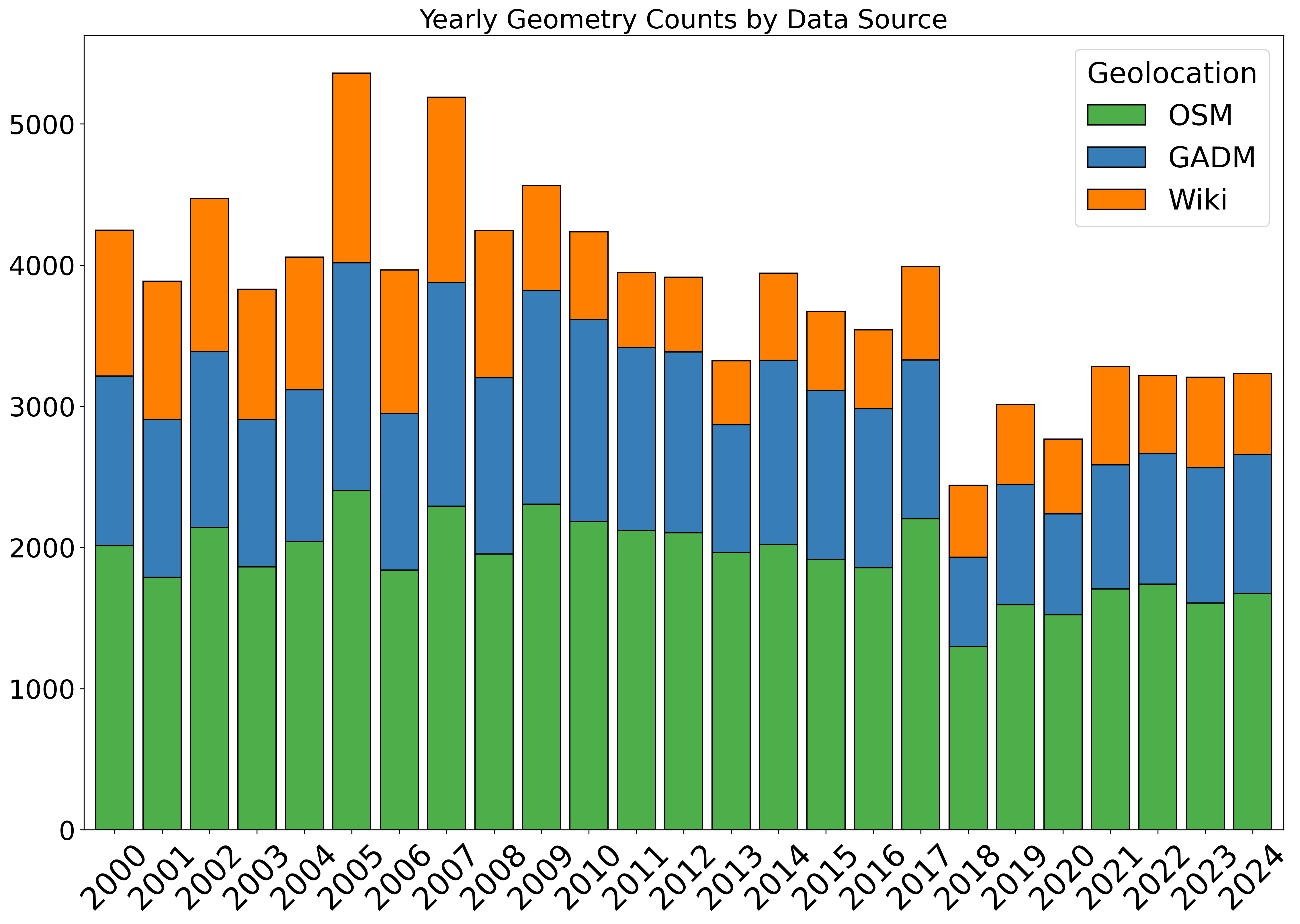}
\caption{This stacked bar plot illustrates the availability of geocoded data for disaster events over time, categorized by data source: OSM, Wikidata, and GADM. Each bar represents the total number of entries per year that have location information from one or more of these sources. The plot highlights the extent to which each source contributes to the coverage of disaster records.}
\label{fig:yearly}
\end{figure}

\begin{figure}[h!]
\centering
\includegraphics[width=15cm,height=8cm]{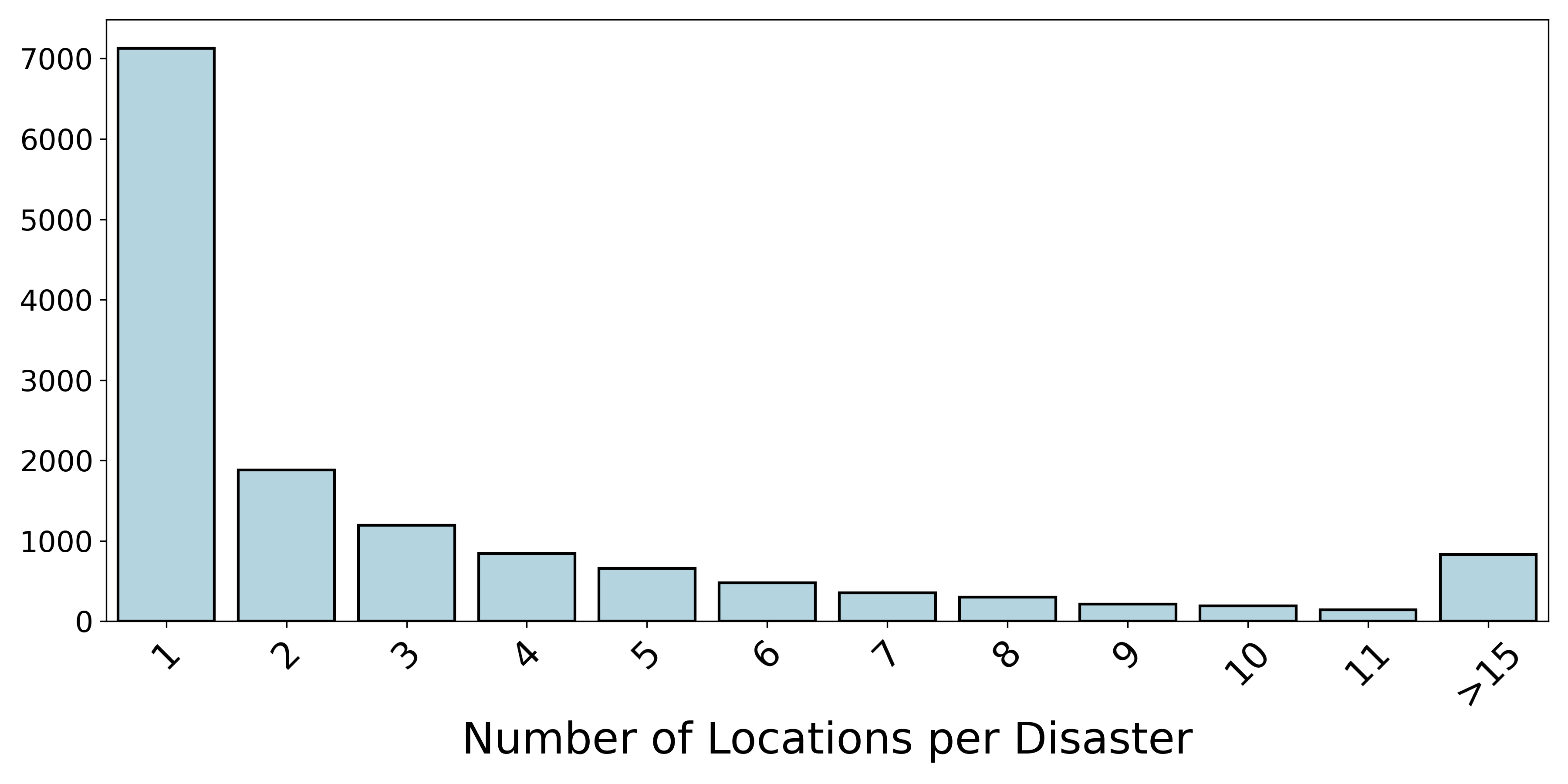}
\caption{Distribution of the number of locations per disaster. The x-axis represents the number of locations, while the y-axis indicates the frequency of disasters with a given number of locations. Note that the histogram has been capped at a number of locations greater than 15.
}
\label{fig:lochist}
\end{figure}

The yearly evolution of the absolute number of geocoded records is shown in Supplementary Figure \ref{fig:yearly}, where bar colors indicate the geocoding source. The relative contribution of each source varies slightly over time, but OSM consistently provides the largest share of matches, followed by GADM and Wikidata. Supplementary Figure \ref{fig:lochist} shows the distribution of the number of locations per disaster, which is right-skewed toward a single-location mode—most disasters affect only one location—while a long tail extends beyond 200 locations for some large-scale events. On average, each disaster is associated with approximately four distinct locations.

\section{Disaster group distribution}

\begin{figure}[h!]
\centering
\includegraphics[width=15cm,height=9cm]{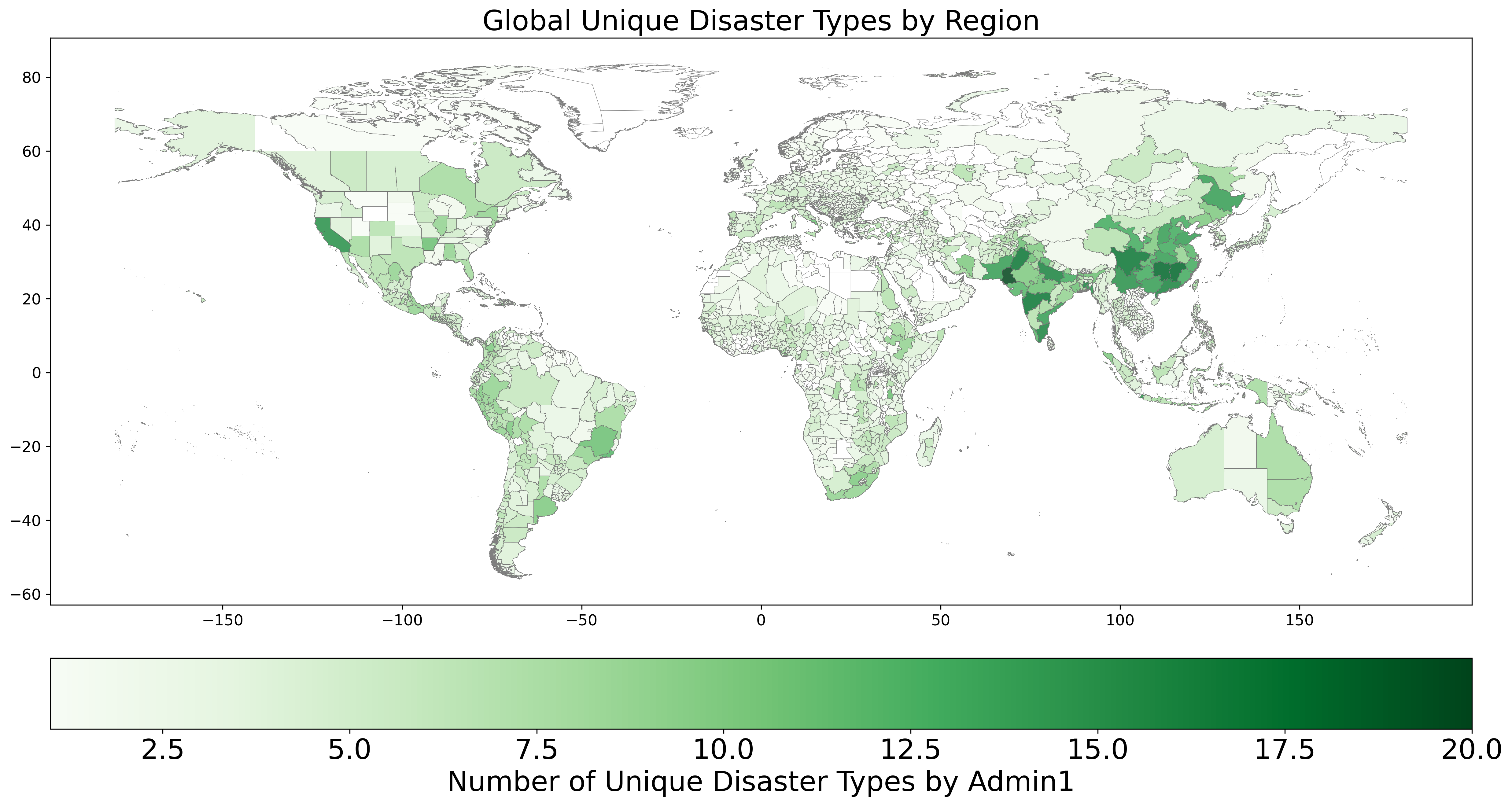}
\caption{Map illustrating the diversity of disaster types across global regions at \texttt{Admin1} level. Each area is shaded according to the number of unique disaster types experienced, ranging from 1 to 31. The color gradient, shown in shades of green, indicates regions with a greater variety of disaster types.
}
\label{fig:map1}
\end{figure}

\begin{figure}[h!]
\centering
\includegraphics[width=12cm,height=18cm]{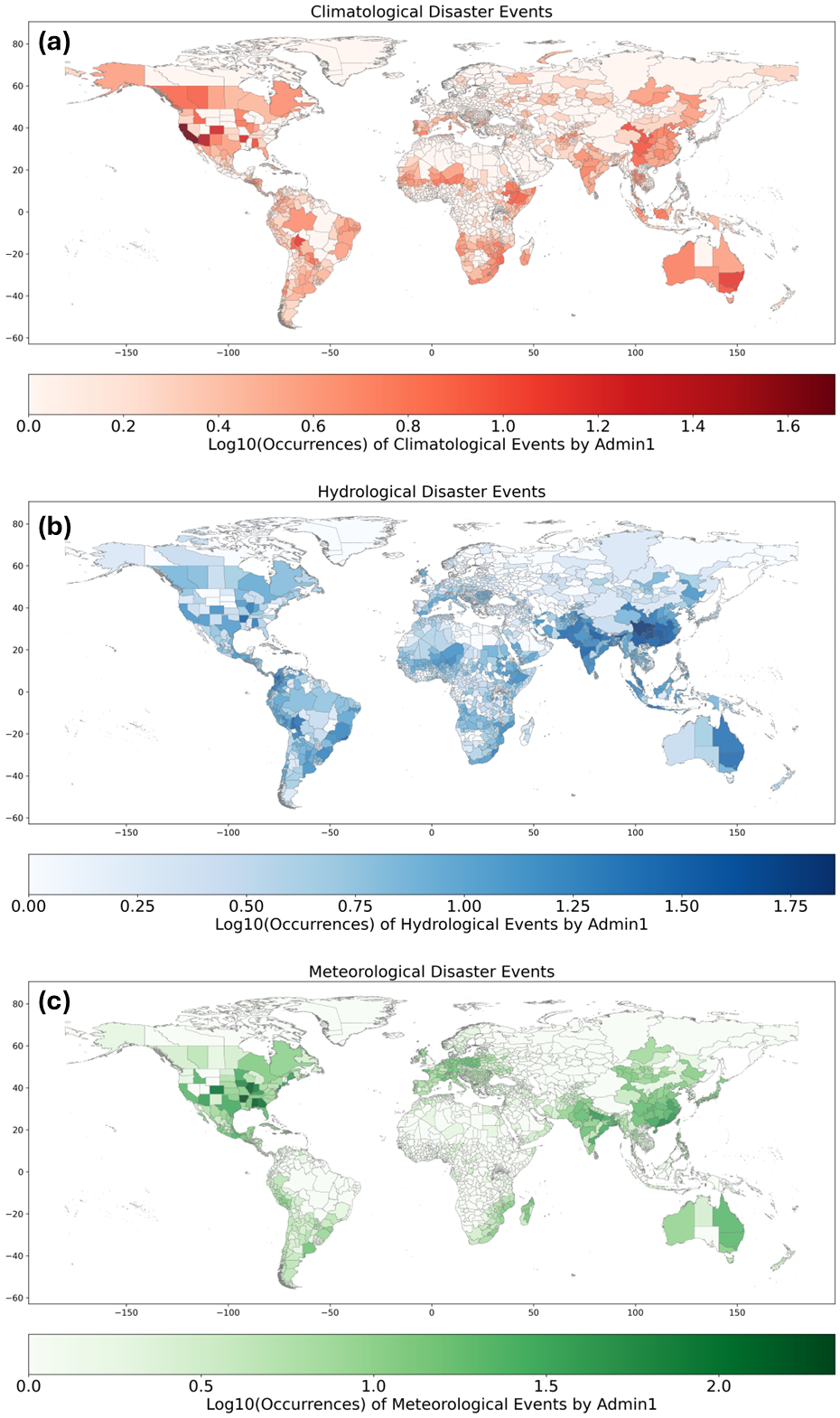}
\caption{Map illustrating the frequency of Climatological (a), Hydrological (b), and Meteorological (c) disaster events at \texttt{Admin1} level. 
}
\label{fig:map2}
\end{figure}

\begin{figure}[h!]
\centering
\includegraphics[width=12cm,height=18cm]{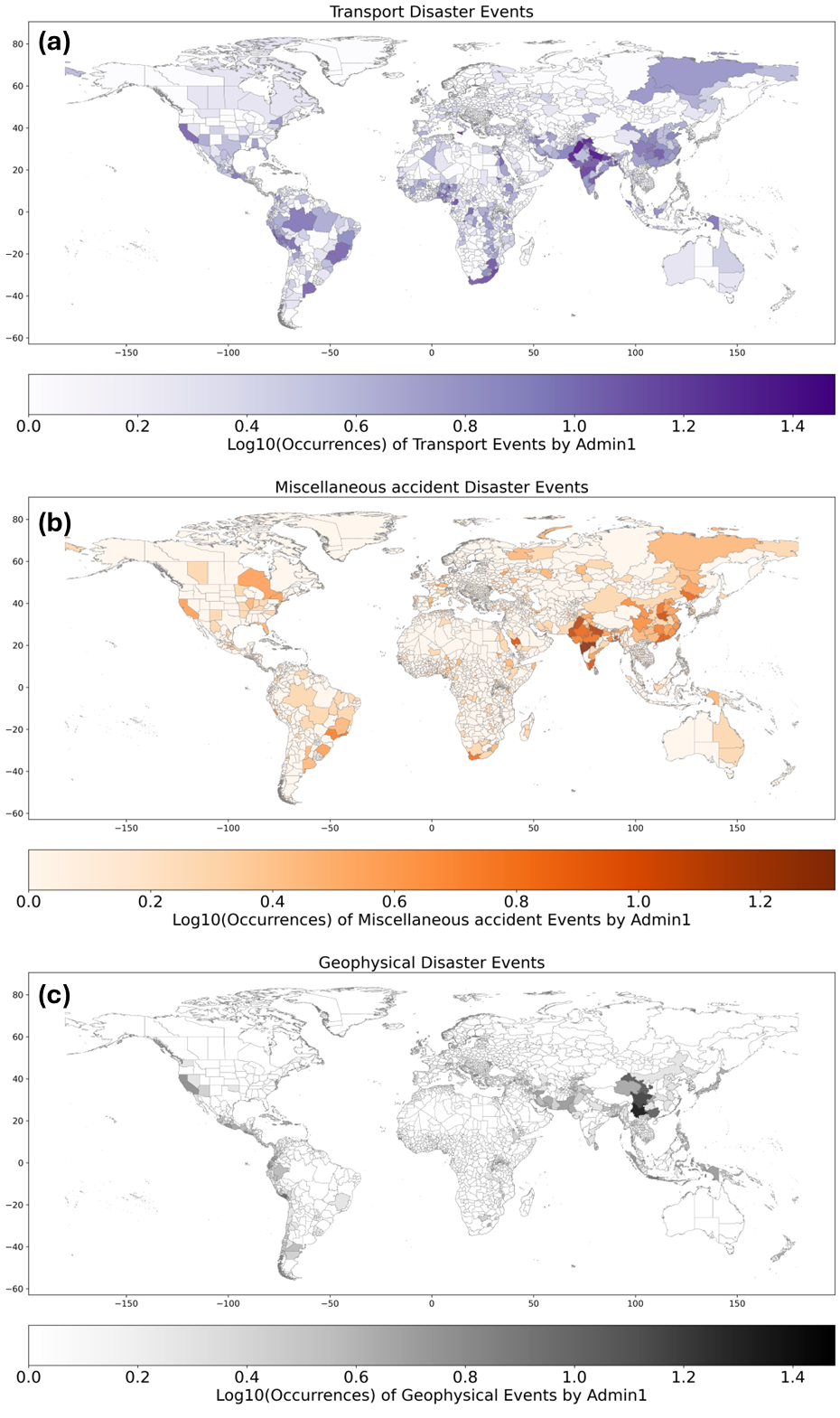}
\caption{Map illustrating the frequency of Transport (a), Miscellaneous accident (b), and Geophysical (c) disaster events at \texttt{Admin1} level. }
\label{fig:map3}
\end{figure}

\begin{figure}[h!]
\centering
\includegraphics[width=12cm,height=18cm]{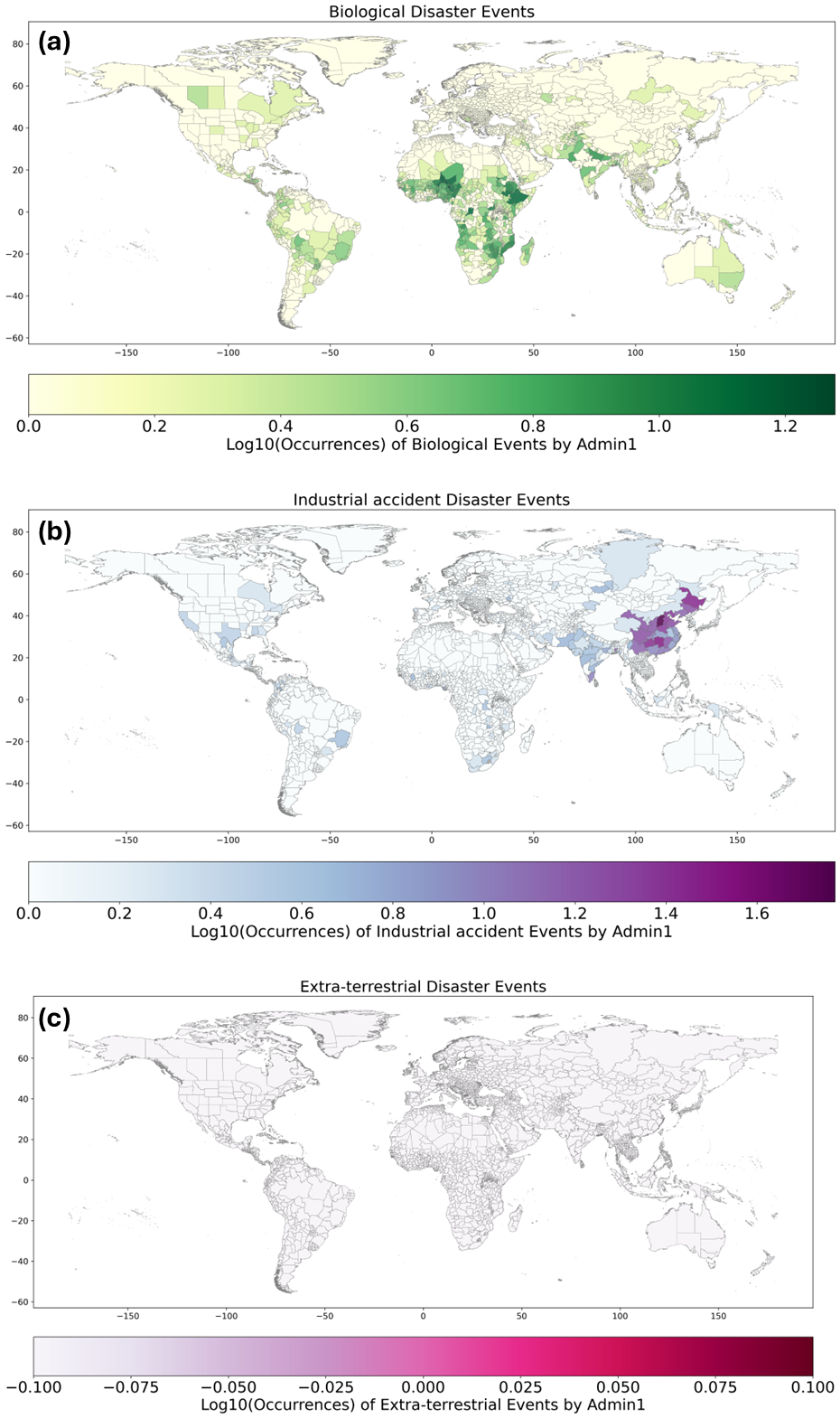}
\caption{Map illustrating the frequency of Biological (a), Industrial (b), and Extra-terrestrial (c) disaster events at \texttt{Admin1} level. }
\label{fig:map4}
\end{figure}

All analyses are conducted and visualized at the GADM \texttt{Admin1} level for consistency across regions. We also map the subnational distribution and diversity of disasters to explore how different hazard types and their combinations vary globally. Supplementary Figure \ref{fig:map1} depicts the diversity of disaster types across regions, where each area is shaded according to the number of distinct disaster types recorded between 2000 and 2024. Darker green tones indicate regions exposed to a broader variety of hazards. See also Figures~\ref{fig:map2}–\ref{fig:map4} for maps showing the (log-transformed) frequency of events by disaster subgroup—including climatological, hydrological, meteorological, transport, miscellaneous accident, geophysical, biological, industrial, and extra-terrestrial disasters. Darker tones of each colorscale correspond to a higher frequency of disasters for the specific group represented.

\section{Reliability distribution}

\begin{figure}[h!]
\centering
\includegraphics[width=12cm,height=10cm]{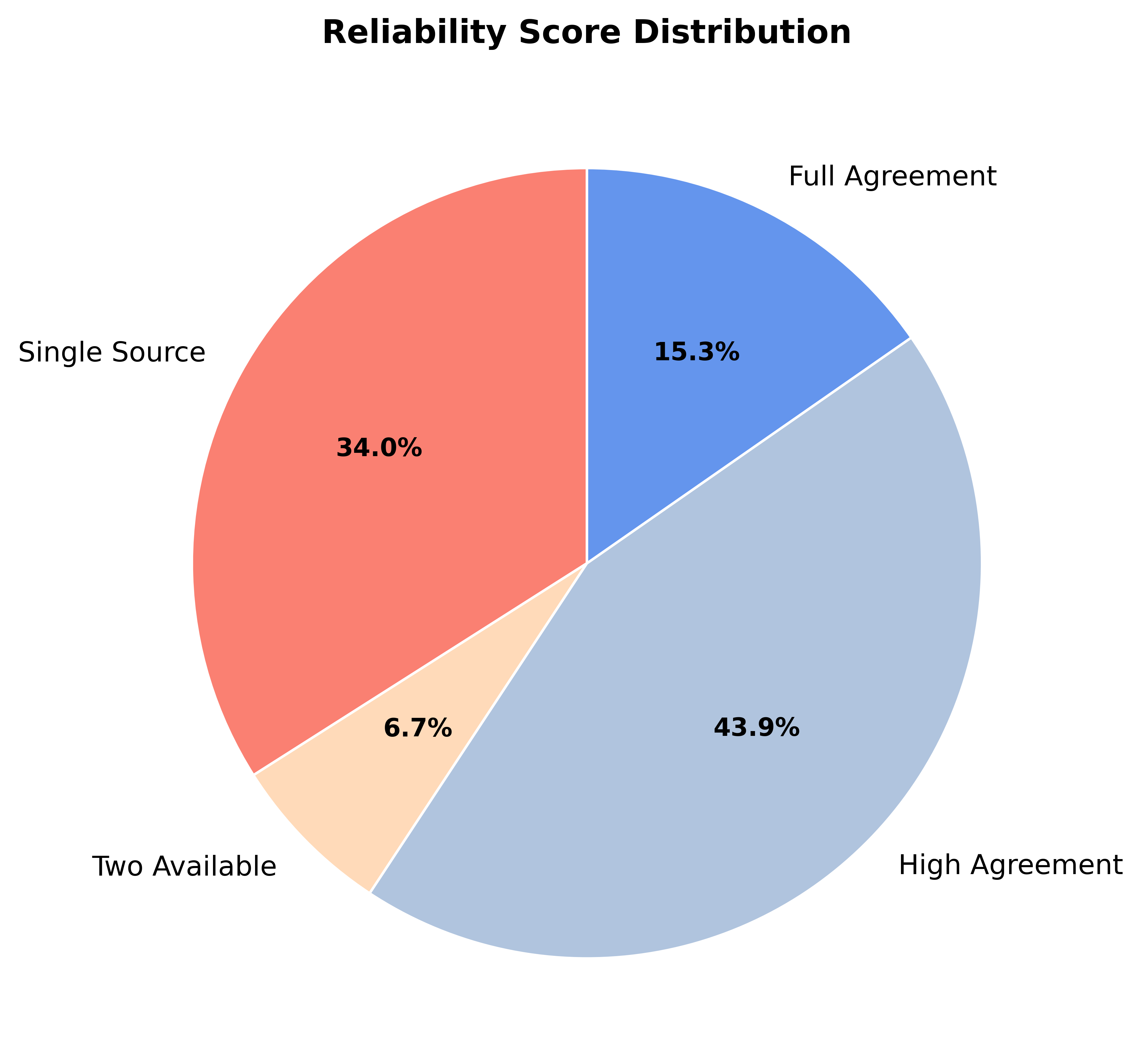}
\caption{This pie chart depicts the distribution of matching reliability scores for available geocoded data, categorized by agreement levels among data sources: Single Source, Two Available, High Agreement, and Full Agreement. 
}
\label{fig:relpie}
\end{figure}

\begin{figure}[h!]
\centering
\includegraphics[width=16cm,height=8cm]{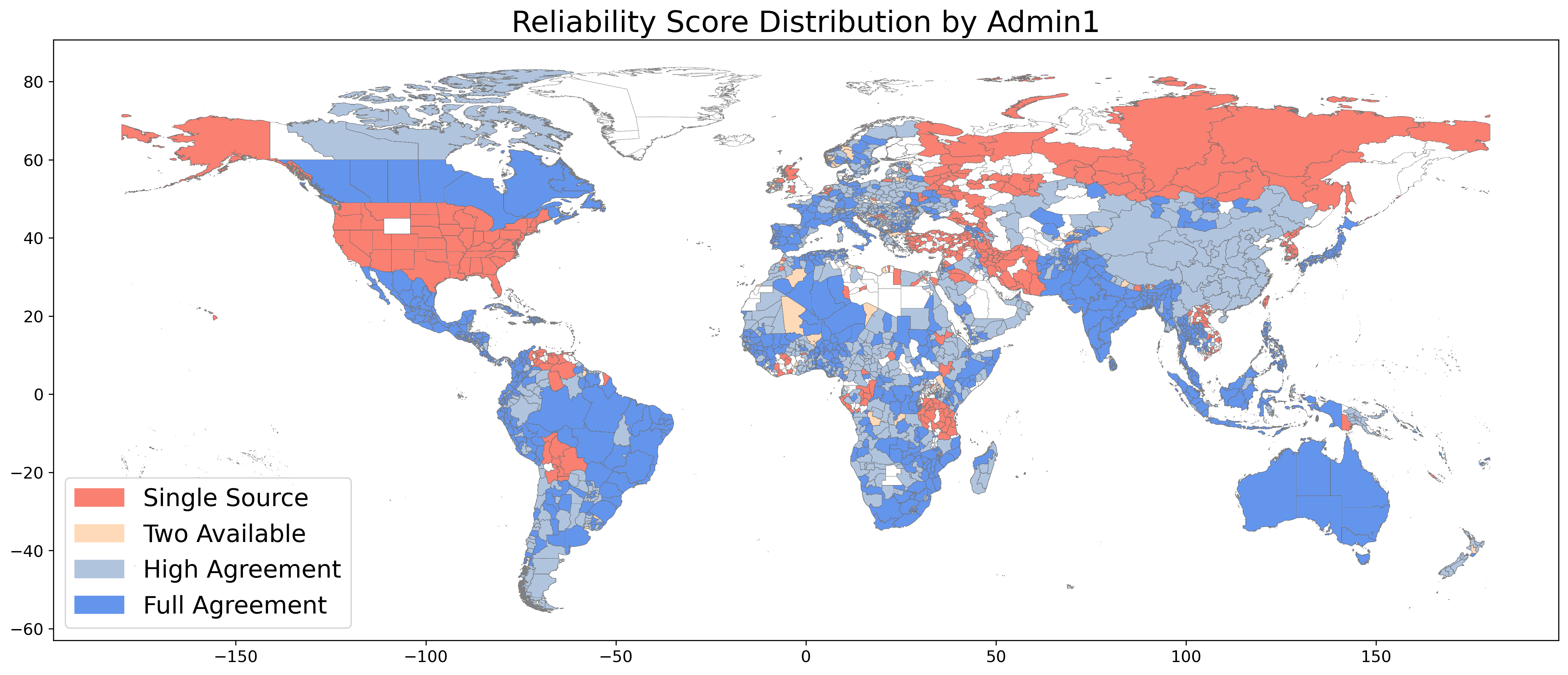}
\caption{Map of matching reliability scores for available geocoded data at \texttt{Admin1} level, categorized by agreement levels among data sources: Single Source, Two Available, High Agreement, and Full Agreement.}
\label{fig:map5}
\end{figure}

We additionally introduce a reliability score that quantifies the consistency of geocding sources for each disaster record. This score reflects whether a location was identified by one, two, or all three sources (OSM, GADM, and Wikidata), and whether these sources agree spatially. See Figure~\ref{fig:relpie} for the overall distribution of reliability categories—Single Source, Two Available, High Agreement, and Full Agreement—and Figure~\ref{fig:map5} for their spatial distribution at the \texttt{Admin1} level, which highlights regional differences in source availability and agreement.

\end{document}